\documentclass[sigconf]{acmart}

\AtBeginDocument{%
  }

\setcopyright{acmcopyright}
\copyrightyear{2023}
\acmYear{2023}
\acmDOI{XXXXXXX.XXXXXXX}

\acmConference[MM '23]{Make sure to enter the correct
  conference title from your rights confirmation email}{October 29--November 03,
  2023}{Ottawa, Canada}
\acmPrice{15.00}
\acmISBN{978-1-4503-XXXX-X/18/06}

\acmSubmissionID{2063}

\usepackage{times}
\usepackage{soul}
\usepackage{url}
\usepackage[utf8]{inputenc}
\usepackage{caption}
\usepackage{amsthm}
\usepackage{booktabs}
\usepackage{algorithm}
\usepackage{algorithmic}
\usepackage[switch]{lineno}
\usepackage{subcaption}
\usepackage{makecell}
\usepackage{stfloats}
\usepackage{amsmath}
\usepackage{booktabs}
\usepackage[inline]{enumitem}
\usepackage{CJKutf8}
\usepackage{comment}
\usepackage{graphicx}
\usepackage{subcaption}

\begin{document}

\setcopyright{none}
\settopmatter{printacmref=false} 
\renewcommand\footnotetextcopyrightpermission[1]{} 
\pagestyle{plain}

\title[PaCaNet]{PaCaNet: A Study on CycleGAN with Transfer Learning for Diversifying Fused Chinese Painting and Calligraphy}

\author{Zuhao Yang}
\authornote{Equal contribution}
\email{yang0756@e.ntu.edu.sg}
\affiliation{
\institution{School of Computer Science and Engineering \\ Nanyang Technological University}
\country{Singapore}
}
\author{Huajun Bai}
\authornotemark[1]
\email{hb364@cornell.edu}
\affiliation{
\institution{Seafog AI}
\country{China}
}

\author{Zhang Luo}
\email{luozhang@seafogai.com}
\affiliation{
\institution{Seafog AI}
\country{China}
}

\author{Yang Xu}
\email{yang.xu@sdsu.edu}
\affiliation{
\institution{Department of Computer Science \\ San Diego State University }
\country{USA}
}

\author{Wei Pang}
\email{w.pang@hw.ac.uk}
\affiliation{
\institution{School of Mathematical and Computer Sciences \\ Heriot-Watt University}
\country{UK}
}

\author{Yue Wang}
\email{wang.yue.f07@kyoto-u.jp}
\affiliation{
\institution{Graduate School of Advanced Integrated Studies in Human Survivability \\ Kyoto University}
\country{Japan}
}

\author{Yisheng Yuan}
\email{yyuan01@risd.edu}
\affiliation{
\institution{Graphic Design Department \\ Rhode Island School of Design}
\country{USA}
}

\author{Yeqi Hu}
\email{huyeqi@stu.ouc.edu.cn}
\affiliation{
\institution{College of Information Science and Engineering \\ Ocean University of China}
\country{China}
}

\author{Yingfang Yuan}
\authornote{Corresponding author}
\email{y.yuan@hw.ac.uk}
\affiliation{
\institution{School of Mathematical and Computer Sciences \\ Heriot-Watt University}
\country{UK}
}

\renewcommand{\shortauthors}{Zuhao Yang et al.}
\begin{abstract}
AI-Generated Content (AIGC) has recently gained a surge in popularity, powered by its high efficiency and consistency in production, and its capability of being customized and diversified. The cross-modality nature of the representation learning mechanism in most AIGC technology allows for more freedom and flexibility in exploring new types of art that would be impossible in the past. Inspired by the pictogram subset of Chinese characters, we proposed PaCaNet, a CycleGAN-based pipeline for producing novel artworks that fuse two different art types, traditional Chinese \emph{painting} and \emph{calligraphy}. In an effort to produce stable and diversified output, we adopted three main technical innovations: 
\begin{enumerate*}
    \item Using one-shot learning to increase the creativity of pre-trained models and diversify the content of the fused images. 
    \item Controlling the preference over generated Chinese calligraphy by freezing randomly sampled parameters in pre-trained models. 
    \item Using a regularization method to encourage the models to produce images similar to Chinese paintings. 
\end{enumerate*} 
Furthermore, we conducted a systematic study to explore the performance of PaCaNet in diversifying fused Chinese painting and calligraphy, which showed satisfying results. In conclusion, we provide a new direction of creating arts by fusing the visual information in paintings and the stroke features in Chinese calligraphy. Our approach creates a unique aesthetic experience rooted in the origination of Chinese hieroglyph characters. It is also a unique opportunity to delve deeper into traditional artwork and, in doing so, to create a meaningful impact on preserving and revitalizing traditional heritage. 
\end{abstract}

\begin{teaserfigure}
    \begin{tabular}{c@{\hspace{1em}}c@{\hspace{1em}}c@{\hspace{1em}}c@{\hspace{1em}}c@{\hspace{1em}}c}
        \includegraphics[width=0.15\textwidth]{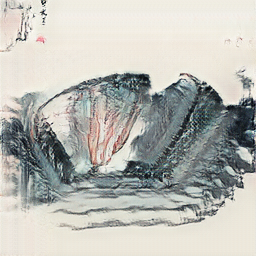} &
        \includegraphics[width=0.15\textwidth]{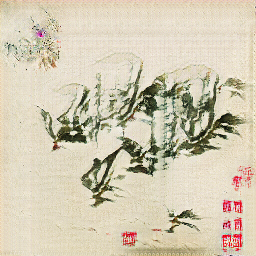} &
        \includegraphics[width=0.15\textwidth]{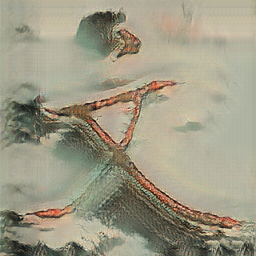} &
        \includegraphics[width=0.15\textwidth]{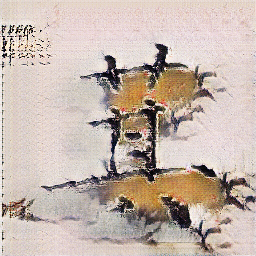} &
        \includegraphics[width=0.15\textwidth]{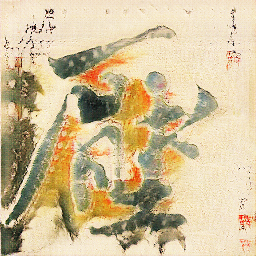} &
        \includegraphics[width=0.15\textwidth]{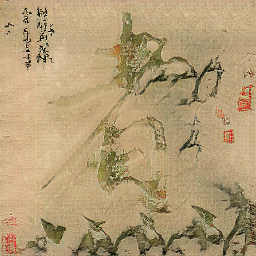} \\[1em]
        \includegraphics[width=0.15\textwidth]{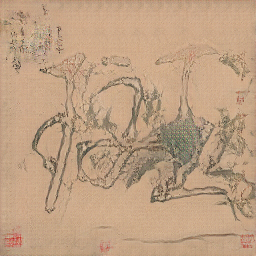} &
        \includegraphics[width=0.15\textwidth]{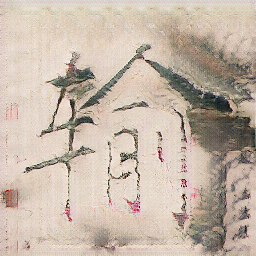} &
        \includegraphics[width=0.15\textwidth]{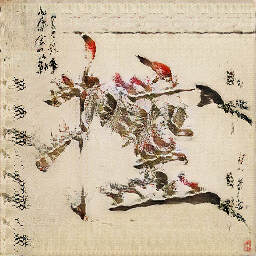} &
        \includegraphics[width=0.15\textwidth]{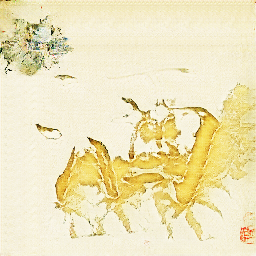} &
        \includegraphics[width=0.15\textwidth]{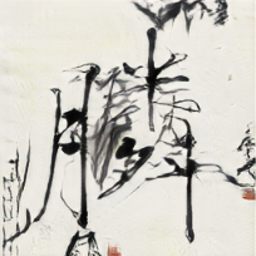} &
        \includegraphics[width=0.15\textwidth]{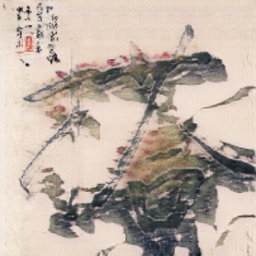}
    \end{tabular}
    \caption{Generated artistic calligraphy fused with various Chinese paintings via our PaCaNet (pre-training stage) framework.}
    \Description{PaCaNet teaser.}
    \label{fig:teaser}
\end{teaserfigure}


\maketitle

\section{Introduction}

AI-generated content (AIGC) refers to all types of content (e.g., text, images, videos, and audio) that are created using artificial intelligence algorithms rather than being produced by humans \cite{pataranutaporn2021ai}. AIGC can be used for a variety of purposes, such as creating stories \cite{roemmele2016writing}, creating posts on social media \cite{radford2019language}, generating literature reviews \cite{aydin2022openai}, and producing creative advertisements \cite{terziouglu2022ad}. People have been investigating AIGC ever since it emerged as a technology. Over the years, research and development in this field have grown significantly. The reason why AIGC is becoming increasingly prevalent is that AI can efficiently produce interesting and creative content based on human knowledge. However, the level of sophistication and quality of AI-generated content can vary greatly depending on the selected algorithms and the real scenarios applied. Our study focuses on diversifying fused Chinese ink painting (we will use the Chinese painting for brevity in the following paragraphs) and Chinese calligraphy.

Chinese painting is a classic form of Chinese art that has been practiced in China for thousands of years. Chinese paintings are well-known for their unique aesthetics and techniques. Over time, the art has evolved to encompass a variety of subjects, such as landscapes, figures, and animals. Chinese painting is also renowned for employing symbolism and metaphor to convey spiritual and philosophical ideals. 

Chinese painting and Chinese calligraphy share similar stylistic elements and tools. The brushstrokes used in paintings are also seen in calligraphy, highlighting the close relationship between the two art forms. However, Chinese calligraphy, which adopts a unique pictographic writing system, has received widespread attention due to its beauty and elegance \cite{xing2016romanization, pengcheng2017chinese}. Through continuous use for thousands of years, this system with such ordering of the various parts and harmony of proportions that the experienced and knowledgeable people will recognize such composition as a work of art \cite{pinder-wilson_barbour_williams_2023}. Unlike the alphabetic writing system, Chinese calligraphy requires to achieve almost perfect proportions within specific grid spaces. Specifically, the structure of Chinese calligraphy is an important aspect that separates it from paintings. Various strokes that are arranged in a particular order form different Chinese characters. The character's structure is crucial since it can alter the character's meaning. Therefore, Chinese calligraphy could provide core guidance on composing specific spatial layouts. Another important characteristic of Chinese calligraphy is the use of rhythm and movement, which is shown through brushstrokes. This can help convey the mood and atmosphere of the scene.

As mentioned above, Chinese painting and calligraphy share many similarities, yet they also possess distinctive characteristics. It would be intriguing to create images that combine the structure of Chinese calligraphy with the features of Chinese painting, conceptually similar to how hieroglyphs represent both a character and an image. As shown in Fig.~\ref{pre-training}, the fused image on the right side retains the structure and brushstrokes of the Chinese calligraphy character from its adjacent image on the left side. At the same time, the fused image also learns the pictorial quality of Chinese painting. 

\begin{figure}
    \centering
    \includegraphics[width=0.45\textwidth]{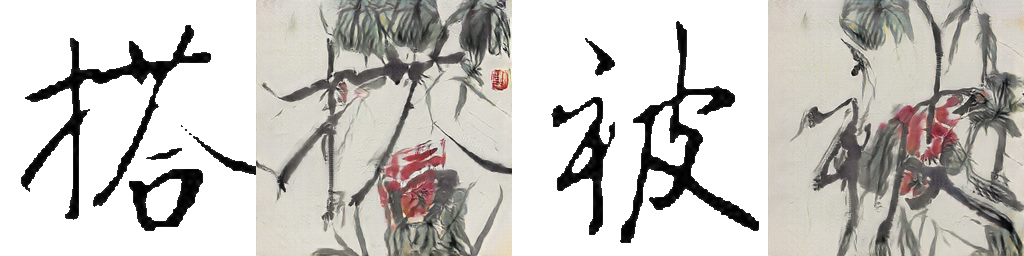}
    \caption{Output images generated by a pre-trained CycleGAN model. Only a small proportion of features are fused.}
    \label{fig:original_cycleGAN}
\end{figure}

In our research, we proposed a new approach, termed Chinese Painting and Calligraphy Network (PaCaNet), to diversify images that fuse the structure of Chinese characters from calligraphy with diverse elements of Chinese paintings. PaCaNet is based on Cycle-Consistent Adversarial Networks (CycleGAN) \cite{CycleGAN_2017} which was originally designed for image-to-image translations \cite{I2I_2016}. The core idea of CycleGAN is to learn two respective distributions to support mapping between two domains. In our experiments, we discovered that the original design of CycleGAN had limitations in generating a diverse range of images. The majority of the generated images were similar in style, which would have a detrimental effect on the originality of AI-generated fused paintings and calligraphy. As shown in Fig.~\ref{fig:teaser} and Fig.~\ref{fig:original_cycleGAN}, we leveraged thousands of distinct unpaired Chinese paintings and calligraphy characters to train the model. However, the model is so restricted that it can only produce similar images that incorporate only a small number of features (e.g. leaf and flower) from thousands of different others present in the training set. This goes against the key criterion of creativity in AIGC \cite{tuomi2023ai}. Furthermore, training a CycleGAN consumes a lot of computational resources, and thus it is not cost-effective to generate a batch of similar images with respect to AIGC. Therefore, we proposed PaCaNet to resolve this dilemma. 

In our study, PaCaNet starts with a source task that is to pre-train a CycleGAN model using unpaired calligraphy images (domain $A$) and Chinese landscape paintings (domain $B$). Domain $A$ contains traditional paintings of many subjects, for example, mountains, flowers, and rivers, but animals are excluded. Once the training process is completed, the generator $G_A$ can map the image $a \in A$ to $B$ while $G_B$ can map the image $b \in B$ to $A$, and $G_A$ can be used to generate fused images (Fig.~\ref{pre-training_figure}). We suppose that $G_A$ and $G_B$ are capable of performing feature representation for images from $A$ and $B$, respectively. Then, we employ three techniques: one-shot learning, parameter freezing, and regularization to improve the creativity of the fused images. \textbf{Firstly}, we diversify the fused images by adding a target task to learn a new mapping from $A$ to $B^{\prime}$ (Chinese animal paintings). In this way, the style of fused images will change since more elements from $B^{\prime}$ will be added. Animal and landscape paintings are different, but both belong to Chinese painting, which means that shared characteristics can be processed by pre-trained $G_A$, and exclusive characteristics can be learned by one-shot learning with a tiny computational budget. By sampling different animal images $b \in B^{\prime}$ as target tasks, we can cost-effectively obtain various-style generators capable of producing diversified fused images for arbitrary calligraphic input. \textbf{Secondly}, to preserve the structure of calligraphy characters in the fused images, we proposed an approach that freezes parameters of the pre-trained model based on random sampling. \textbf{Thirdly}, we introduced a regularization method to enhance the capability of the generator $G^{\prime}_A$ to produce images similar to $b \in B^{\prime}$. In fact, $B^{\prime}$ can be generalized to other types of images without limitation to animals. A more detailed description of PaCaNet will be given in Section \ref{sec:pacanet}. Furthermore, we conducted a systematic study to explore the performance of PaCaNet, and the results will be discussed in Section \ref{sec:exp}.

In summary, the main contributions of our work are as follows:
\begin{itemize}
    \item We proposed PaCaNet as a pioneering framework for exploring the combination of two traditional Chinese cultural heritages. The generated images combine the structural features of Chinese characters and the pictorial features of Chinese paintings.
    \item We collected and organized public datasets for Chinese painting and calligraphy. Our curated dataset will be released shortly to support other research.
    \item We confirmed that freezing randomly sampled parameters performs better than layer freezing when performing transfer learning.
    \item We introduced a regularization method that enables PaCaNet to learn from the specific target, and thus improved the diversity of output.
\end{itemize}

\section{Related Work}
\subsection{Cycle-Consistent Adversarial Networks}
Cycle-Consistent Adversarial Networks (CycleGAN) is a type of generative model that is capable of performing image-to-image translations without the need for paired training data. The CycleGAN generators are trained using a cycle consistency loss, which ensures that the translated image can be transformed back to its original domain while still being similar to the original image. There are several applications that emerge afterward, such as Cycle-MedGAN \cite{MedGAN_2019} for medical image translation, CycleGAN for music genre transfer \cite{MusicGAN_2018}, and CycleGAN for the generation of handwritten Chinese characters \cite{Character_2018}, etc. From our viewpoint, even if CycleGAN has achieved promising performance on various tasks, it still lacks exploration on AIGC.

\subsection{AIGC for Chinese Painting and Calligraphy}
AIGC is the application of machine learning techniques to generate creative content. It includes using deep learning models such as Generative Adversarial Networks (GANs) \cite{GAN_2014} or Latent Diffusion Model (LDM) \cite{SD_2021} to produce new works of art. LDM has been widely explored and proved to be a state-of-the-art model in generating creative and aesthetic images from prompt tuning. However, applying LDM to our task may lead to some semantic information loss, for example, original structure of the calligraphic characters may be discarded \cite{SD_2021}. According to our literature review, we noticed that an equally important yet underexplored problem is how to fuse the artistic style of Chinese paintings with Chinese calligraphy. 
Most of the earliest ancient Chinese characters are \emph{pictograms} (\begin{CJK}{UTF8}{gbsn}象形文\end{CJK}), which are highly stylized and simplified pictures of objects. This unique property makes pictogram characters perfect candidates for creating visual arts, because it is natural to consider them as an abstract representation that contains partial visual information of the object being denoted, which can potentially have similar patterns with paintings of the same object. Related to this idea is the fact that raw visual information in Chinese characters and glyph have been proven useful in language modeling \cite{shi2015radical,meng2019glyce}, but we believe their potentials in AIGC have not been explored. 
One of the major challenges in AIGC for fused Chinese painting and calligraphy is the high level of complexity and variation in the style and technique of traditional Chinese art. We overcome this by using a large selected dataset, which includes various types of high-quality traditional Chinese paintings (e.g., animals, landscapes, and flowers). The AIGC of the traditional Chinese cultural heritage is still worthy of our study.

\subsection{Transfer Learning}
Transfer learning is a technique in machine learning in which a model developed for one task is reused as a starting point for another task. The idea is to transfer the knowledge learned from the first task to the second task, which can be useful in cases where the second task has limited data or is related to the first task.

\textbf{One-shot learning} is a type of machine learning algorithm in which the model is trained to assess the similarity and difference between two images based on a single example. \cite{fei2006one} shows rather than learning from scratch, one can take advantage of the knowledge gathered from previously learned source images to learn about a single example in the target domain. \cite{lake2011one} implemented one-shot learning to learn strokes in novel characters from previous knowledge of how handwritten characters are composed from strokes. The one-shot learning model outperforms a competing state-of-the-art character model and provides a good fit to human perceptual data. 

\textbf{Layer freezing} in transfer learning refers to the process of keeping certain parameters of network layers of a pre-trained model fixed while allowing other parameters to be updated during training. It helps prevent the model from forgetting the features it has learned during the initial pre-training phase. In the study of malware classification \cite{rezende2017malicious}, a model consisting of ResNet-50 layers \cite{resnet_2016} pre-trained on ImageNet \cite{imagenet_2009} is trained by freezing the first 49 layers and leaving the last layer to be trainable. It achieves high accuracy and shows that the knowledge obtained in the ImageNet classification task can be successfully transferred to malware classification tasks. In addition, adaptive transfer learning approaches are proposed to better fine-tune the pre-trained model on the new task. For example, SpotTune \cite{guo2019spottune} automatically decides the optimal set of layers, and stochastic depth \cite{huang2016deep} proposes a linearly decayed survival probability to drop a subset of layers in residual blocks \cite{resnet_2016}.

\section{Method} \label{sec:pacanet}
\begin{figure*}
    \centering
    \includegraphics[scale=0.58]{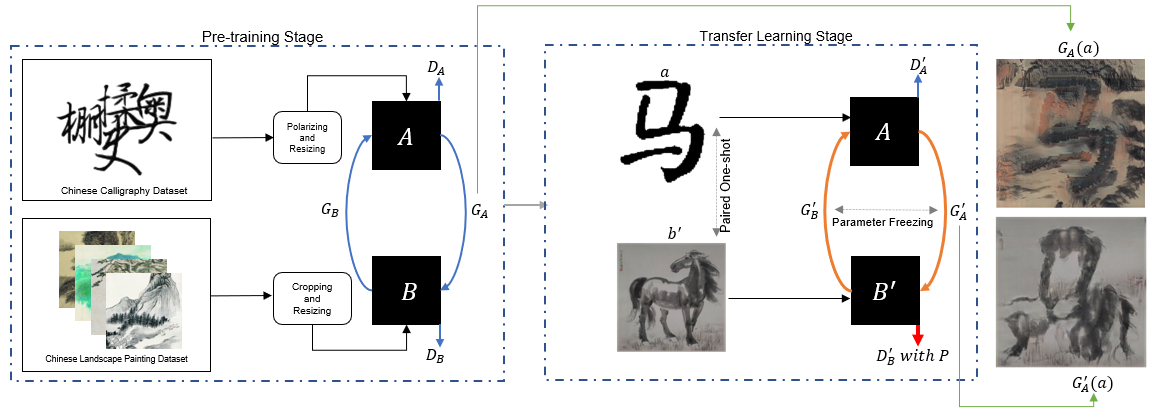}
    \caption{Overall framework of PaCaNet. The red arrow in the transfer learning stage indicates that the discriminator $D^{\prime}_B$ is regularized with a penalty term $P$. Green arrows indicate that the images $G_A(a)$ and $G^{\prime}_A(a)$ are respectively generated by generators $G_A$ and $G^{\prime}_A$. Orange arrows denote that parameters of $G^{\prime}_A$ and $G^{\prime}_B$ are partially frozen in our one-shot learning method during the transfer learning stage.}
    \label{fig:pacanet_framework}
\end{figure*}

The proposed PaCaNet framework is shown in Fig.~\ref{fig:pacanet_framework}. $A$ and $B$ respectively represent two domains for Chinese landscape painting and Chinese calligraphy. $G_A$ and $G_B$ denote the generators for two mappings $G_A$: $A \rightarrow B$ and $G_B$: $B \rightarrow A$. In general, PaCaNet consists of two stages. The first stage, called the pre-training stage, aims to produce a model that processes landscape paintings and calligraphy. Before training, landscape images will be cropped and resized, as the sizes of most Chinese painting images are large and vary widely, which causes an overwhelming computational cost. Meanwhile, the calligraphy dataset will be resized and polarized. Further details of polarization will be discussed in our experimental part (Section \ref{sec: datasets}).

During the second stage (transfer learning stage), the user is allowed to add a preferred pair of images (e.g., an arbitrary Chinese animal painting $b^{\prime}$ with any Chinese calligraphy character $a$) to perform \textbf{one-shot learning} on the pre-trained model obtained from the previous stage, which involves a new generator $G_A^{\prime}$ for mapping $A \rightarrow B^\prime$ where $B^\prime$ denotes the domain of Chinese animal paintings.

It is essential to keep in mind that during the transfer learning phase, the discriminators $D_A$ and $D_B$ will be updated to $D^{\prime}_A$ and $D^{\prime}_B$, with all their parameters being modified accordingly. Meanwhile, $G_A$ and $G_B$ will be \textbf{partially frozen} to become $G_A^{\prime}$ and $G_B^{\prime}$. To encourage $G_A^{\prime}$ to learn more features from $b^{\prime} \in B^\prime$, we proposed a \textbf{regularization} method that adds a penalty term $P$ to the mapping $A \rightarrow B^\prime$. Once transfer learning has been completed, $G_A^{\prime}$ can generate images that combine features of $B^\prime$ with arbitrary calligraphic, producing more diversified outputs than those of original $G_A$.

In the following paragraphs, the use of transfer learning and regularization will be discussed in detail.


\subsection{Transfer Learning for Learning Features of Calligraphy} \label{sec:tl}
The original CycleGAN can generate fused images, but with limited creativity for different calligraphic input, as shown in Fig.~\ref{fig:original_cycleGAN}. 
Towards the goal of introducing more 
creativity (i.e., generating images with distinctive artistic styles for different Chinese characters as input), we applied two transfer learning techniques to our pre-trained CycleGAN model, which are: \textbf{one-shot learning} and \textbf{parameter freezing}.

One-shot learning involves a pair of Chinese calligraphy character $a$ and Chinese animal painting $b'$. Both as datasets for Chinese paintings, $B$ mostly includes landscapes, while $B'$ contains only animals. Our pre-trained CycleGAN model has generalization abilities for Chinese paintings. Adding one-shot learning will develop its ability to integrate features of animal paintings. One-shot learning occurs when $G_A^{\prime}$ learns how to map $a$ to $b^{\prime}$, with other network architectures of CycleGAN $G_B^{\prime}$, $D_A^{\prime}$ and $D_B^{\prime}$  iterating normally as the training stage. The learning process only requires one pair of ($a,b'$) to fine-tune the target task. In the end, we will obtain a generator $G_A^{\prime}$ that is capable of generating the fused image $G_A^{\prime}(a)$ with both the features from Chinese landscape paintings $B$ those from the animal image $b'$.

However, we discovered that calligraphic characteristics and landscape characteristics of $B$ are lost with one-shot learning. 
Parameter freezing is a solution we adopted in transfer learning stage to make the model produce better calligraphy handwriting and retain the features learned in the pre-training stage. One common way is to freeze the parameters by layers \cite{imagenet_2009}: freeze all the parameters of one ResNet block of the generator network (details in Section \ref{sec:pf}). 
We found a better way to freeze the parameters by randomly sampling all network parameters with a freezing rate $r \in [0, 1]$, as illustrated in Algorithm \ref{alg:algorithm}. A freezing rate of $r=0.9$ means that 90\% of all parameters of the generator networks will not be trainable during the one-shot learning stage. The fused images generated in this way are as good as those generated by freezing layers. It also offers users an adjustable freezing rate $r$ to control the level of features learned in the pre-training stage and the character remembrance for the output images.

\begin{algorithm}[tb]
    \caption{Parameter freezing algorithm of generators}
    \label{alg:algorithm}
    \makebox[\linewidth][l]{
        \begin{tabular}{@{}l@{\hspace{1mm}}l}
            \multicolumn{2}{@{}l}{\textbf{Input}: Generators $G_A$ and $G_B$.} \\
            \multicolumn{2}{@{}l}{\textbf{Hyperparameter}: Freezing rate $r_f \in [0, 1]$.} \\
            \multicolumn{2}{@{}l}{\textbf{Output}: Generators $G_A^{\prime}$ and $G_B^{\prime}$.}
        \end{tabular}
    }
    \begin{algorithmic}[1] 
        \FOR{ $g \text{ in } \{G_A^{\prime}, G_B^{\prime}\}$}  
        \FORALL{parameters $p  \text{ in }  g.parameters()$}
        \STATE Generate a random number $r \in [0, 1]$.
        \IF {$r < r_f$}
        \STATE Freeze parameter $p$.
        \ELSE
        \STATE Make parameter $p$ trainable.
        \ENDIF
        \ENDFOR
        \ENDFOR
    \end{algorithmic}
\end{algorithm}

\subsection{Regularization for Learning Features of Paintings} \label{sec:reg}
CycleGAN aims to learn mapping functions between $A$ and $B$ given unpaired training data. The original loss function of CycleGAN is proposed as Eq.~\ref{original_cyclegan}:
\begin{equation}
    \begin{aligned}
        \mathcal{L}\left(G_A, G_B, D_B, D_A\right) & =\mathcal{L}_{\mathrm{GAN}}\left(G_A, D_B, A, B\right) \\
        & +\mathcal{L}_{\mathrm{GAN}}\left(G_B, D_A, B, A\right) \\
        & + \lambda\mathcal{L}_{\mathrm{cyc}}\left(G_A, G_B\right),
    \end{aligned}
    \label{original_cyclegan}
\end{equation}
where $\mathcal{L}\textsubscript{GAN}$ stands for adversarial loss \cite{GAN_2014}; $\mathcal{L}\textsubscript{cyc}$ represents cycle consistency loss \cite{CycleGAN_2017}; and $\lambda$ is a coefficient that controls the relative weighting of the above two objectives.

We expect the generated images to combine Chinese calligraphy with traditional Chinese paintings. The transfer learning approaches mentioned in Section \ref{sec:tl} have already fulfilled the need to fuse calligraphy characters. However, according to our experiments, we discovered that CycleGAN with transfer learning still underperformed in learning high-level features from $B^\prime$.

To ameliorate this issue, we introduced a penalty term $P$ applied to the original CycleGAN. Initially, we chose Root Mean Square Error (RMSE) and Multi-scale Structural Similarity Index Measure (MS-SSIM) \cite{msssim_2003} to calculate the pixel-level similarity between $G_A^{\prime}(a)$ and $b^\prime$ for an arbitrary image $a \in A$ and its paired image $b^\prime \in B^\prime$ during the one-shot learning process. Empirically, we found that using RMSE only actually reconstructed better $B^\prime$-like images that preserved most of the features and looked extremely similar to the image $b$ we used during the transfer learning stage. Nevertheless, what we desire is that the fused images exhibit as many features from Chinese paintings as possible without completely losing the handwriting of the calligraphy characters. Consequently, we only adopted $MS-SSIM$ as a penalty term $P$ that is used to define Eq.~\ref{reg_term}:

\begin{equation}
    \mathcal{L}\textsubscript{reg}(G_A^{\prime}, B^\prime) = -(MS-SSIM(G_A^{\prime}(a), b^\prime) + C),
    \label{reg_term}
\end{equation}
where $\mathcal{L}\textsubscript{reg}$ denotes the loss produced by the penalty term; $C$ denotes a constant term that is empirically set to 1 to avoid a negative loss value.

Here, $\mathcal{L}\textsubscript{reg}$ is solely related to $D_B^{\prime}$ because we only focus on improving the discriminator ability that distinguishes between the fused image and its paired image from $B^\prime$. $\mathcal{L}\textsubscript{reg}$ plays a critical role in driving the model to learn more characteristics from $B^{\prime}$. Thus, the full objective during the transfer learning stage becomes:
\begin{equation}
    \begin{aligned}
        \mathcal{L}\left(G_A^{\prime}, G_B^{\prime}, D_B^{\prime}, D_A^{\prime}\right) & =\mathcal{L}_{\mathrm{GAN}}\left(G_A^{\prime}, D_B^{\prime}, A, B^{\prime}\right) \\
        & +\mathcal{L}_{\mathrm{GAN}}\left(G_B^{\prime}, D_A^{\prime}, B^{\prime}, A\right) \\
        & + \lambda_{1}\mathcal{L}_{\mathrm{cyc}}\left(G_A^{\prime}, G_B^{\prime}\right) \\
        & + \lambda_{2}\mathcal{L}\textsubscript{reg}\left(G_A^{\prime}, B^\prime\right),
    \end{aligned}
\end{equation}
where $\lambda_1$ and $\lambda_2$ are two hyperparameters that control the relative importance of $\mathcal{L}_{cyc}$ and $\mathcal{L}_{reg}$, respectively. Our aim is to find out:
\begin{equation}
    {G^\prime_A}^*, {G^\prime_B}^* = \arg\underset{G_A^{\prime}, G_B^{\prime}} {\min} \underset{D_B^{\prime}, D_A^{\prime}} {\max} \mathcal{L}(G_A^{\prime}, G_B^{\prime}, D_B^{\prime}, D_A^{\prime}).
\end{equation}

\section{Experimental Results}\label{sec:exp}
\subsection{Datasets and Input Pre-processing}\label{sec: datasets}
Our experiments strictly followed the workflow presented in Fig.~\ref{fig:pacanet_framework}. We initially collected 40,000 landscape paintings of several Chinese artists (e.g., Baishi Qi, Beihong Xu, and Binhong Huang). Furthermore, we selected 4,745 of them as domain $B$ by manual evaluation. We also downloaded 4,745 Chinese calligraphy images from \cite{kaggle_2020} as domain $A$ to make up our own unpaired dataset for the CycleGAN model, called PaCa dataset. 80\% data of the PaCa dataset is used to pre-train the model with 200 epochs. In addition, we prepared 100 animal paintings as domain $B^{\prime}$ for transfer learning.

\begin{figure}[!htbp]
    \centering
    \begin{subfigure}[h]{0.225\textwidth}
        \includegraphics[width=\textwidth]{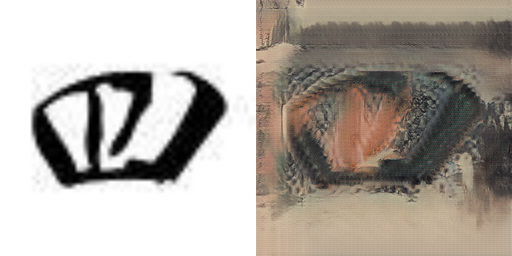}
        \caption{An example of pre-training result given unpolarized input. (Left): Original calligraphy. (Right): Fused image.}
    \label{pre-training_nonploar}
    \end{subfigure}
    \hspace{1em}
    \begin{subfigure}[h]{0.225\textwidth}
        \includegraphics[width=\textwidth]{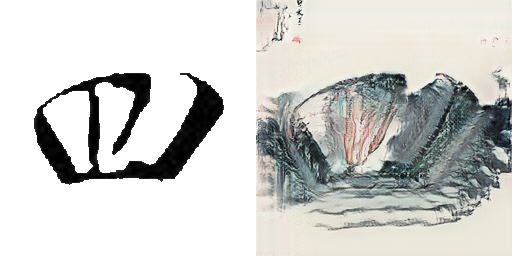}
        \caption{An example of pre-training result given polarized input. (Left): Original calligraphy. (Right): Fused image.}
        \label{pre-training}
    \end{subfigure}
    \caption{Pre-training results given unpolarized (left) and polarized (right) input.}
    \label{pre-training_figure}
\end{figure}

During the pre-processing phase, we resized every image to $256\times256$ pixels by pixels. Moreover, we polarized our calligraphic input so that every pixel intensity value turns into either 0 or 255 (in grayscale). Intuitively, the handwriting should be pure black, and the background should be entirely white. Empirically, we found that the polarization removed all the noise around the handwriting (shown in the left half of Fig.~\ref{pre-training_nonploar} and Fig.~\ref{pre-training}). Meanwhile, we noticed that the fused images without polarization showed a much worse effect (Fig.~\ref{pre-training_nonploar}). In contrast, it can be seen in Fig.~\ref{pre-training} that some features, such as the calligraphic inscription and the ripple effect of the water, are well preserved.


\subsection{Comparison Experiment on Naïve CycleGAN and CycleGAN with One-shot Learning} \label{sec:osl}
In the following experiments, we set naïve CycleGAN as our baseline model and explore how one-shot learning upon our pre-trained model contributes to the creativity and diversity of CycleGAN. We started our study with a Chinese ``horse'' painting. The hyperparameters are set as follows: the number of epochs with initial learning rate ($\alpha=0.0002$) was 100 and the number of epochs with linearly decayed learning rate was 100; the Adam optimizer \cite{adam_2014} was adopted for training. Convergence occurred around the $50^{th}$ epoch of the learning rate decay phase.

\begin{figure}[!htbp]
    \centering
    \includegraphics[width=0.45\textwidth]{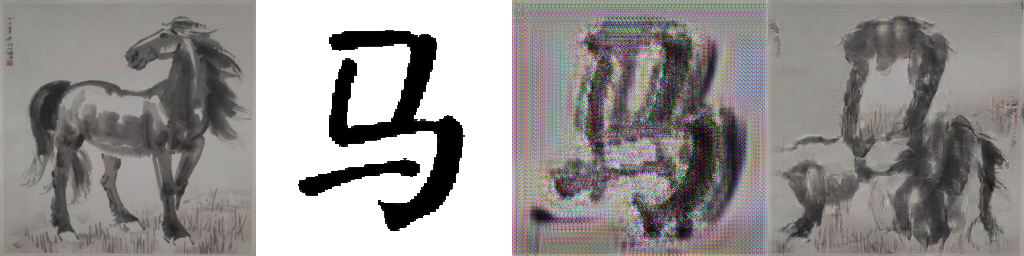}
    \caption{Synthesis results of simplified Chinese calligraphy ``horse'' character ($2^{nd}$ image) with horse painting (leftmost) generated by naïve CycleGAN ($3^{rd}$ image) and CycleGAN with one-shot learning (rightmost). Here, \begin{CJK}{UTF8}{gbsn}``马''\end{CJK} is the Chinese character corresponding to ``horse''.}
    \label{naive_simplified}  
\end{figure}

We observe in Fig.~\ref{naive_simplified} that the simplified Chinese calligraphy character of ``horse'' fails to produce a decent result of style fusion. Gird-like noise is produced in the fused image. In addition, we can barely see the features learned from the original horse painting. In this paper, we applied the one-shot learning method that uses only one pair of training images (e.g., calligraphy character of ``horse'' and traditional horse painting) to adjust the fusion details over the images generated by our pre-trained model.

\begin{figure*}[t]
    \centering
    \begin{subfigure}[b]{0.45\textwidth}
      \begin{tabular}{c|cc|cc|}
        {\includegraphics[width=0.195\linewidth]{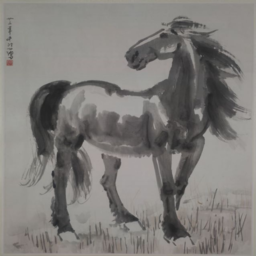}} &
        {\includegraphics[width=0.195\linewidth]{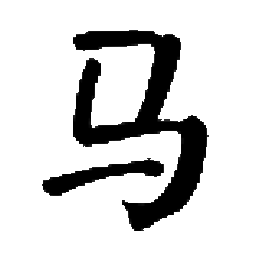}} &
        {\includegraphics[width=0.195\linewidth]{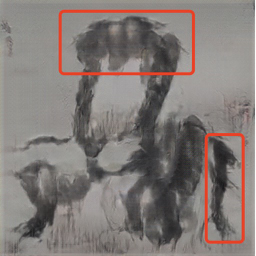}} &
        {\includegraphics[width=0.195\linewidth]{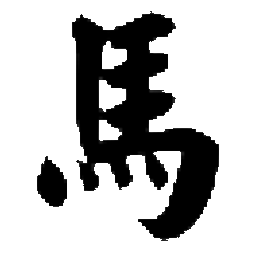}} &
        {\includegraphics[width=0.195\linewidth]{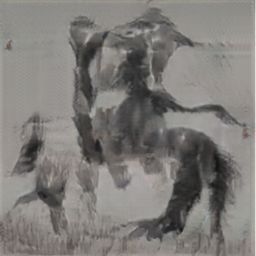}} \\
        {\includegraphics[width=0.195\linewidth]{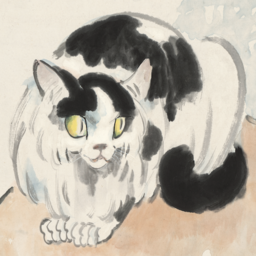}} &
        {\includegraphics[width=0.195\linewidth]{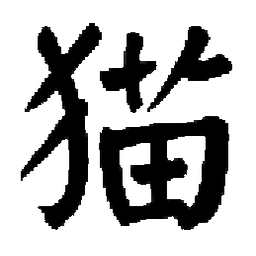}} &
        {\includegraphics[width=0.195\linewidth]{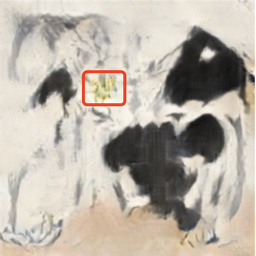}} &
        {\includegraphics[width=0.195\linewidth]{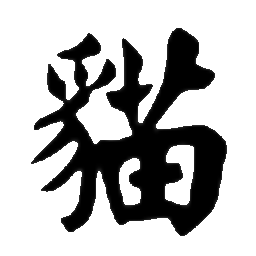}} &
        {\includegraphics[width=0.195\linewidth]{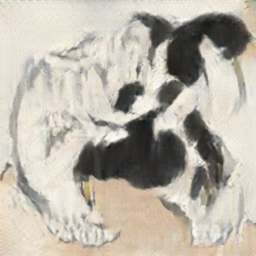}} \\
        {\includegraphics[width=0.195\linewidth]{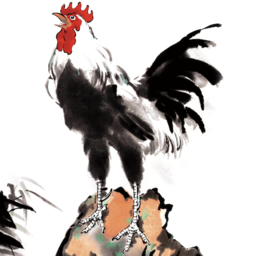}} &
        {\includegraphics[width=0.195\linewidth]{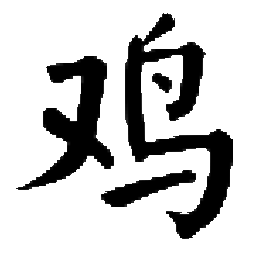}} &
        {\includegraphics[width=0.195\linewidth]{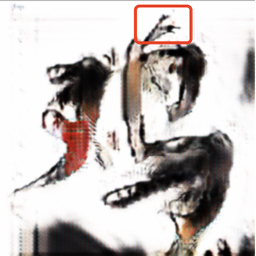}} &
        {\includegraphics[width=0.195\linewidth]{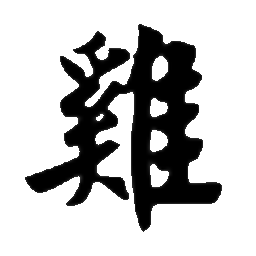}} &
        {\includegraphics[width=0.195\linewidth]{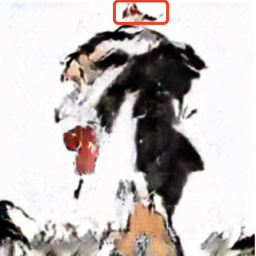}} \\
        {\includegraphics[width=0.195\linewidth]{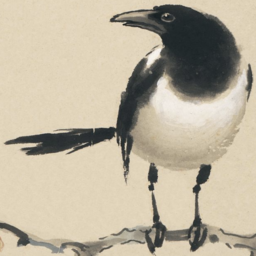}} &
        {\includegraphics[width=0.195\linewidth]{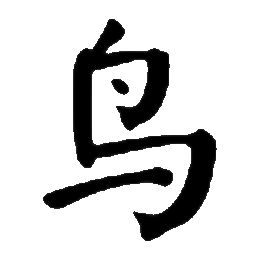}} &
        {\includegraphics[width=0.195\linewidth]{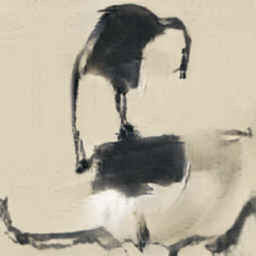}} &
        {\includegraphics[width=0.195\linewidth]{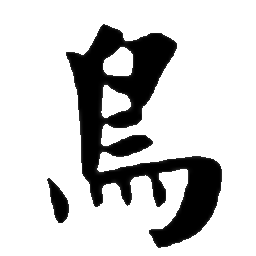}} &
        {\includegraphics[width=0.195\linewidth]{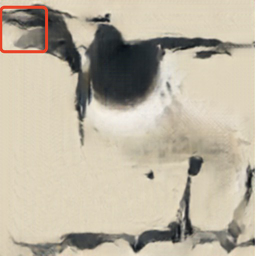}} \\
      \end{tabular}
      \subcaption{Synthesis results of simplified and traditional Chinese calligraphy characters generated by CycleGAN with one-shot learning approach. (Left): Animal paintings. (Middle): Simplified Chinese calligraphy character and corresponding fused image. (Right): Traditional Chinese calligraphy character and corresponding fused image.}
      \label{fsl_example}
    \end{subfigure}
    \hfill
    \begin{subfigure}[b]{0.45\textwidth}
      \begin{tabular}{cc|cc}
        {\includegraphics[width=0.195\linewidth]{images/horse_char_simp.png}} &
        {\includegraphics[width=0.195\linewidth]{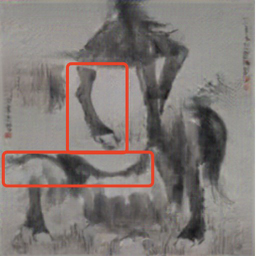}} &
        {\includegraphics[width=0.195\linewidth]{images/horse_char_trad.png}} &
        {\includegraphics[width=0.195\linewidth]{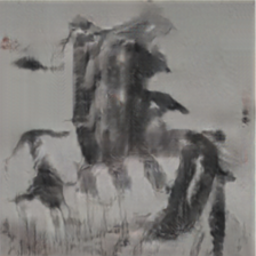}} \\
        {\includegraphics[width=0.195\linewidth]{images/cat_char_simp.png}} &
        {\includegraphics[width=0.195\linewidth]{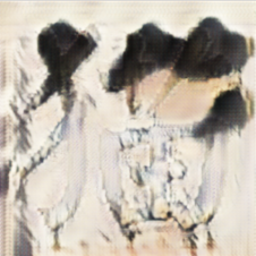}} &
        {\includegraphics[width=0.195\linewidth]{images/cat_char_trad.png}} &
        {\includegraphics[width=0.195\linewidth]{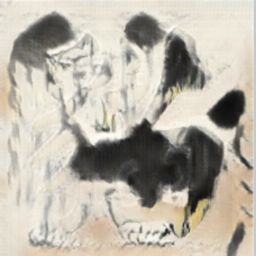}} \\
        {\includegraphics[width=0.195\linewidth]{images/chicken_char_simp.png}} &
        {\includegraphics[width=0.195\linewidth]{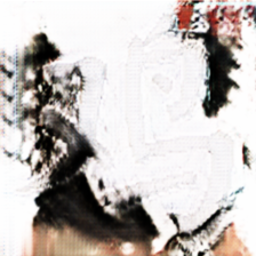}} &
        {\includegraphics[width=0.195\linewidth]{images/chicken_char_trad.png}} &
        {\includegraphics[width=0.195\linewidth]{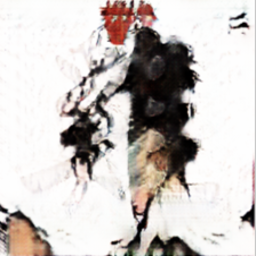}} \\
        {\includegraphics[width=0.195\linewidth]{images/bird_char_simp.png}} &
        {\includegraphics[width=0.195\linewidth]{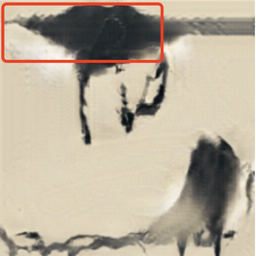}} &
        {\includegraphics[width=0.195\linewidth]{images/bird_char_trad.png}} &
        {\includegraphics[width=0.195\linewidth]{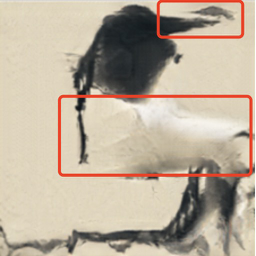}} \\
      \end{tabular}
      \caption{Synthesis results of simplified and traditional Chinese calligraphy characters generated by CycleGAN with one-shot learning approach and regularization. (Left): Simplified Chinese calligraphy character and corresponding fused image. (Right): Traditional Chinese calligraphy character and corresponding fused image.}
      \label{bc_example}
  \end{subfigure}
  \caption{Synthesis results. (Left): CycleGAN with one-shot learning. (Right): CycleGAN with one-shot learning as well as regularization. In our study, we investigated Chinese calligraphy character of animals and paired them with their corresponding animal painting.}
  \label{overview}
\end{figure*}

We extended our case study of the horse to three more animals, i.e., cat (\begin{CJK}{UTF8}{gbsn}``猫''\end{CJK}), rooster (\begin{CJK}{UTF8}{gbsn}``鸡''\end{CJK}), and bird (\begin{CJK}{UTF8}{gbsn}``鸟''\end{CJK}). We performed experiments for both simplified and traditional Chinese calligraphy characters. As can be observed in Fig.~\ref{fsl_example}, the horse's mane, body, and tail are portrayed vividly. When it comes to the cat, its yellow eyes and black spots are captured. With well-preserved original handwriting, the fusion of traditional calligraphy characters gains a better effect than their simplified counterparts. For instance, the traditional ``rooster'' character generates a cockscomb right above the image, in which the cockscomb of the original rooster painting resides. Nevertheless, the fused result of the simplified ``rooster'' only exhibits some feather-like effect. Meanwhile, the traditional ``bird'' character maintains the bird's beak after the transfer learning stage, while its simplified counterpart does not.

Obviously, one-shot learning based on our pre-trained model yields drastically better fusion effects compared to the naïve CycleGAN. This is due to the fact that our baseline model adopts normal weight initialization \cite{normal_init_2010}, whereas our proposed CycleGAN model with the one-shot learning method continues training from a pre-trained model on a training set of Chinese landscape paintings. Nonetheless, the one-shot learning approach tends to overfit the animal paintings, leading to the loss of calligraphic structures and brushstrokes.

\subsection{CycleGAN with One-shot Learning and Parameter Freezing} \label{sec:pf}
\begin{figure}[!htbp]
    \centering
    \includegraphics[scale=0.33] {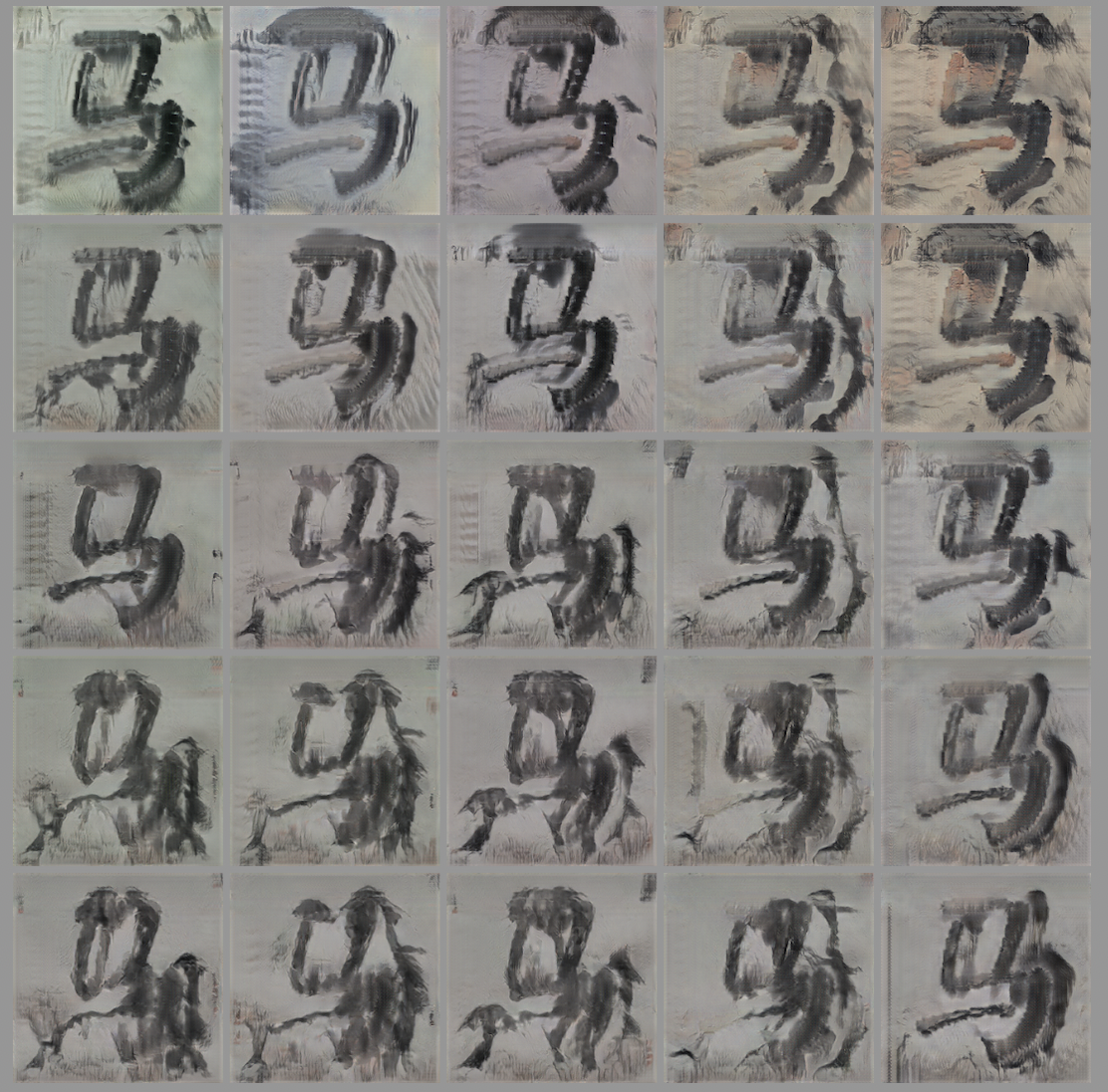} 
    \caption{Generated images through testing on different epochs (vertically) with different freezing rates (horizontally). The number of epochs from top to bottom is 10, 20, 60, 120, 200. Freezing rate from left to right is 0.1, 0.2, 0.5, 0.8, 0.9.}
    \label{fig:grid_arrows}
\end{figure}
\begin{figure}[!htbp]
    \centering
    \includegraphics[scale=0.2]{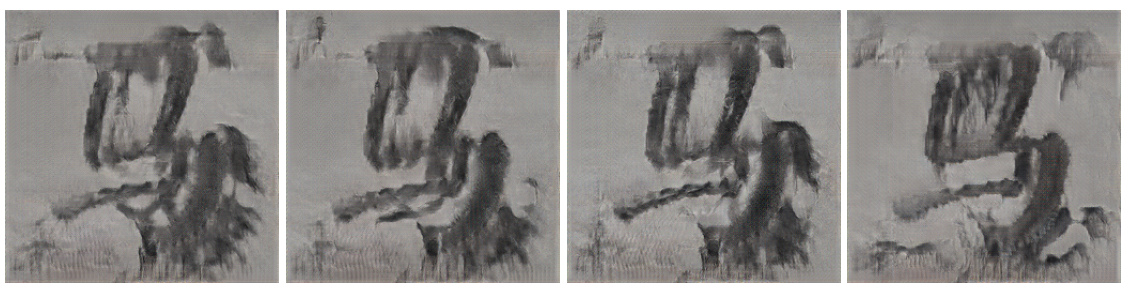}
    \caption{Generated images by freezing different ResNet blocks. From left to right, the figure shows results of freezing only the $3^{rd}$, $5^{th}$, $7^{th}$ and $9^{th}$ unit of 9 ResNet blocks in $G^{\prime}_A$ and $G^{\prime}_B$ of PaCaNet.}
    \label{fig:freeze_blocks}
\end{figure}

From the experimental results in Section \ref{sec:osl}, we note that CycleGAN with one-shot learning can learn representations of the animal task. Yet, it loses information learned in the pre-training stage: the fused images show little pictorial elements of Chinese landscape paintings. Moreover, the structure of the Chinese calligraphy character is weakened during the one-shot learning stage.
To keep the image features learned at both stages, we introduced a method that performs one-shot learning with parameter freezing for the network.

Inspired by the previous study \cite{huang2016deep}, we decided to freeze part of the PaCaNet parameters and leave the rest trainable. The network structure is as follows: the generators $G_A$ and $G_B$ of our network consisting of 2 downsampling layers, 9 ResNet blocks, and 2 upsampling layers, sequentially \cite{Johnson2016Perceptual}. We conducted two sets of experiments as follows. 

In the first set, we investigated the effect of image fusion by freezing one ResNet block but on different indexes of layers. As shown in Fig.~\ref{fig:freeze_blocks}, the results of the four groups look similar. We hypothesized that as long as the total number of freezing parameters remains the same, the results should look alike. Since one ResNet block takes up about 10\% of the total parameters in the entire generator network, we introduced the approach that freezes 10\% of the total parameters as described in 
Section \ref{sec:tl}. Subsequently, we conducted an experiment with the freezing rate $r=0.1$, which produced a similar outcome to the freezing of one ResNet block, which validated our hypothesis.


We further carried out another set of experiments to investigate the effect of changing the freezing rate $r$. As shown in Fig.~\ref{fig:grid_arrows}, a trend of the mixture of features of Chinese landscape paintings and horse painting is displayed.
With fewer epochs or more frozen parameters, the model will remember more features from the pre-trained stage, i.e., features of Chinese landscape paintings, preserving the structure of the calligraphy character of ``horse'' at the same time. Thereby, on one hand, the more towards the upper right corner, the more features of Chinese landscape paintings can be found. On the other hand, with more training epochs or fewer frozen parameters, the model tends to learn more features of the animal paintings. Hence, the farther away the bottom left corner, the more features of the animal painting can be observed.

To summarize, the freezing rate allows us to adjust the level of blending of characteristics from Chinese landscape and animal paintings in the fused image.

\subsection{CycleGAN with One-shot Learning and Regularization}
Parameter freezing approach discussed in Section \ref{sec:pf} addressed the issue of missing calligraphic features. However, it degrades the performance of CycleGAN in learning features from $B^\prime$. Hence, we applied the regularization technique (penalty term) during the transfer learning stage together with the one-shot learning introduced in Section \ref{sec:osl}.

Empirically, we verified that the generated images became more unlike the original handwritten character after regularization. Meanwhile, the fused images managed to learn more features from the animal paintings, as suggested in Section \ref{sec:reg}. Referring to Fig.~\ref{bc_example}, the ``horse''example presents an artificial hieroglyph, in which the handwriting of the original character is further weakened. Besides, the horse's hoof and the lower side of its body take the positions of vertical and horizontal stroke, respectively. 

Using the set of traditional calligraphy characters as input still outperforms its simplified counterpart. For example, the fused bird produced by the simplified ``bird'' character only maintains the bird's head and beak from the target image, whereas its traditional counterpart preserves birds' bodies well on top of that. We believe there are two major aspects that contribute to the better synthesis effect of traditional calligraphy characters. First, most traditional characters possess more complex structures, which provides more features to learn. Second, traditional characters are visually closer to hieroglyphs, which caters to our aesthetic idea of fusing handwritten strokes with Chinese paintings.

\subsection{Ablation Study}
In this section, we will provide a brief summary of the performance of our previous experimental groups. In particular, we will focus on quantitative evaluation and qualitative analysis to distinguish between four models: original CycleGAN (CycleGAN\textsubscript{Naïve}), CycleGAN with one-shot learning (CycleGAN\textsubscript{OSL}), CycleGAN with one-shot learning combined with parameter freezing (CycleGAN\textsubscript{OSL+PF}), and CycleGAN with one-shot learning incorporating both parameter freezing and regularization (CycleGAN\textsubscript{OSL+PF+REG}).

We selected 100 unseen animal paintings and paired them with their corresponding calligraphy image to perform one-shot learning as before. The model was still trained on a 200-epoch basis to produce 100 generators ($G^{\prime}_A$) with distinctive styles. Afterwards, 1,000 unused calligraphy images were leveraged to run inferences on these generators, which consequently produced 100 sets of test results. We respectively computed the Fréchet Inception Distance (FID) \cite{FID_2017} and Fréchet Pre-train Distance (FPD) \cite{FPD_2021} scores of the fused results regarding $A$ and $B^\prime$. Through calculating the average value of these two evaluation metrics over 100 sets of fused images, the overall performance of the original CycleGAN and our proposed CycleGAN variants can be determined.

\begin{table}[!h]
    \centering
    \resizebox{0.45\textwidth}{11.1mm}{
        \begin{tabular}{l r r r r}
            \toprule
            \textbf{Method} & $FID_A$ & $FPD_A$ & $FID_{B^\prime}$ & $FPD_{B^\prime}$ \\
            \midrule
            CycleGAN\textsubscript{Naïve} & \textbf{400.904} & 2.002 & 422.825 & 0.291 \\
            CycleGAN\textsubscript{OSL} & 477.531 & 1.863 & 373.592 & \textbf{0.221} \\
            CycleGAN\textsubscript{OSL+PF} & 458.155 & \textbf{1.775} & 465.507 & 1.812  \\
            \textbf{CycleGAN\textsubscript{OSL+PF+REG}} & 460.299 & 1.794 & \textbf{369.597} & 0.280 \\
            \bottomrule
        \end{tabular}
    }
    \caption{Overall evaluation using quantitative metrics for our CycleGANs (with transfer learning).}
    \label{metric_table}
\end{table}



We aim to generate fused images integrating the characteristics of animals with the structure of calligraphy characters. This can be verified from the quantitative results shown in Table.~\ref{metric_table}. Our proposed model, PaCaNet (CycleGAN\textsubscript{OSL+PF+REG}), performed best in terms of $FID_{B^{\prime}}$ compared to other approaches. It also had similar performance to CycleGAN\textsubscript{Naïve} and CycleGAN\textsubscript{OSL} in terms of $FPD_{B^{\prime}}$, indicating that PaCaNet effectively learned the features from $B^{\prime}$. CycleGAN\textsubscript{Naïve} yielded the best score in terms of $FID_A$, since it generated images that were almost identical to the characters in $A$ without learning any features from $B^{\prime}$. When it comes to $FID_A$ and $FPD_A$, PaCaNet was neither the best nor the worst, which is in line with our expectations as we did not want the generated images to display calligraphy characters only. Overall, while CycleGAN\textsubscript{OSL} performed similarly to PaCaNet on $B^{\prime}$, PaCaNet behaved better with respect to both $FID_A$ and $FPD_A$, making it the top-performing model. 

\begin{figure}[H]
    \centering
    \includegraphics[scale=0.15]{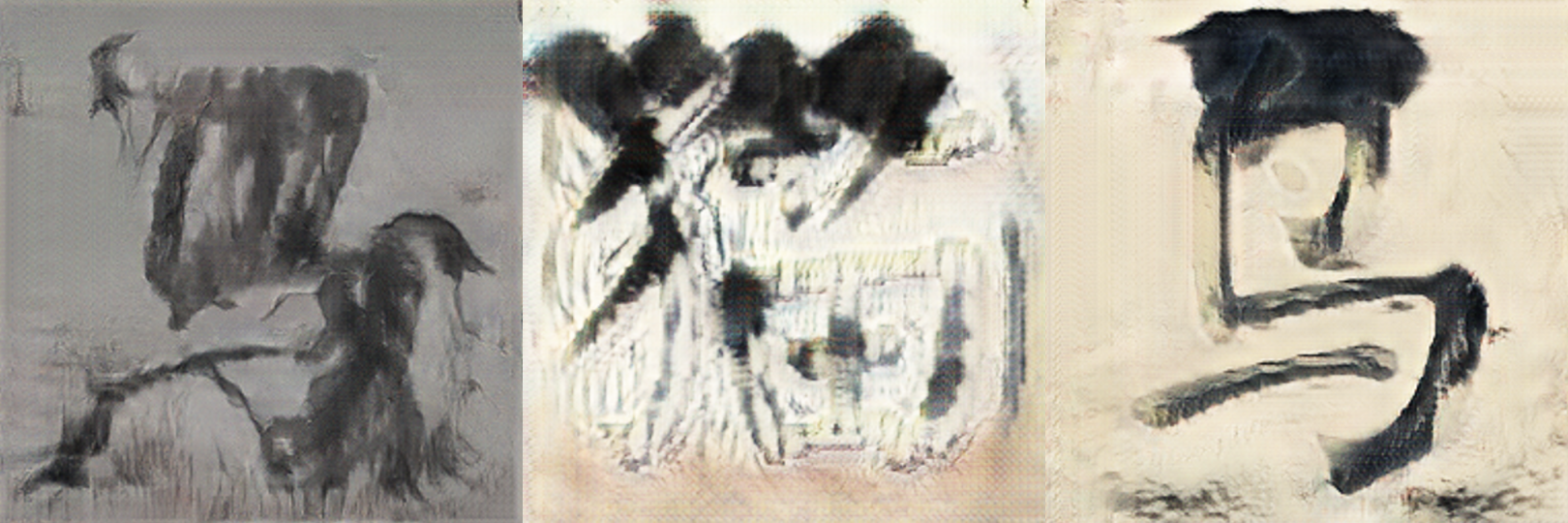}
    \caption{Synthesis results of CycleGAN combined with our proposed parameter freezing and regularization methods. (Left): Horse. (Middle): Cat. (Right): Bird.}
    \label{as_example}
\end{figure}

As shown in Fig.~\ref{as_example}, clearer calligraphic strokes of ``cat'' and ``bird'' characters can be observed compared to the results we obtained using CycleGAN\textsubscript{OSL} (Fig.~\ref{fsl_example}) and CycleGAN\textsubscript{OSL+REG} (Fig.~\ref{bc_example}). Meanwhile, the pictorial elements of $B^\prime$ are presented elegantly.

\subsection{Human Evaluation}
To complete the qualitative analysis of the outputs from PaCaNet. We invited two artists to evaluate the value of the generated images.

\textbf{Reviewer 1}: In this study, by combining animal imagery in Chinese painting with its corresponding Chinese calligraphy, the output results (Fig.~\ref{overview}) demonstrate that simple character structures can help make the picture more compact and show a more stable composition in the given space. Although complex structures may not exhibit the advantages of composition well in the output, by rearranging the distribution of blocks according to the character structure, if AI can learn composition through Chinese calligraphy with a specific purpose, it may be able to output a more stable and harmonious visual structure more effectively in the future.

In addition, incorporating texture can achieve the expression of subjective cultural intentions. Through learning the color blocks and texture of Chinese painting, the output results show a similar atmosphere, which retains the characteristics of Chinese paintings and completes the cultural expression to some extent. Considering the connotation of Chinese pictographic characters, their font structure itself is a conversion of the form and meaning of external objects. In other words, the character structure of Chinese characters has a set of developed logic for decoding external objects. By repeatedly comparing and learning this logic, it may be possible to develop a new visual artistic structural innovation capability that can help non-physical intentions (such as emotions, logic, etc.) to decode with Chinese graphics. This implies that the character structure of Chinese calligraphy is not only a means of conveying written language but also a representation of external objects. The logic behind the formation of Chinese characters can serve as a foundation for developing new and innovative visual art structures that can facilitate the decoding and interpretation of non-physical intentions in Chinese graphics.

\textbf{Reviewer 2}: In my humble opinion, AI is, in the first place, a tool for productivity. The trained AI model described in this paper is capable of mass-calculating and mass-creating an abundance of materials considered the fusion of Chinese calligraphy and ink paintings based on patterns. By that, we see how human control is inserted into artistic production (under loose definition) that takes advantage of existing historical and cultural assets to meet a certain extent of expectations. It requires very little previous formal artistic training. By examining the results closer, for instance, the horse (\begin{CJK}{UTF8}{gbsn}``马''\end{CJK}) and bird (\begin{CJK}{UTF8}{gbsn}``鸟''\end{CJK}) cases in Fig.~\ref{overview}, we can notice how the image-production logic reveals the characteristics of Chinese pictograms (e.g., characters resemble the forms and gestures of the animals depicted). It serves as an entrance to the further study of archeology on signs, philology on writing, and linguistics through the artistic landscapes provided by the paintings. From prehistoric signage to the revolution of simplified Chinese characters, with the assistance of visual art as the pictorial representation, I see the potential of utilizing these models in the aforementioned areas of study to address the logocentrism that still prevails. I would also like to emphasize that, from the above case, AI-generated outcomes can be entry points but not the final product; references but not answers; and guides but not canons.

\section{Conclusions and Future Work}
In this study, we introduce PaCaNet, a cutting-edge transfer learning model designed to generate diverse fusion of Chinese painting and calligraphy. 
The generated output images present new aesthetic effects that have not been created in previous generative models.  
By incorporating one-shot learning, random parameter freezing, and a novel regularization technique, PaCaNet demonstrates superior performance compared to other methods through extensive experiments. Our results showcase the effectiveness of PaCaNet in producing unique and high-quality fused images. Additionally, the artistic style fusion will open a new avenue for exploring the traditional cultural heritage. In the future, we will try to examine the capability of our PaCaNet framework to combine both visual and textual inputs in the image fusion process. Furthermore, we may adapt our approach to other artistic domains, such as \emph{neoclassicism} and \emph{impressionism}.




\bibliographystyle{ACM-Reference-Format}
\bibliography{sample-base}


\begin{thebibliography}{32}


\ifx \showCODEN    \undefined \def \showCODEN     #1{\unskip}     \fi
\ifx \showDOI      \undefined \def \showDOI       #1{#1}\fi
\ifx \showISBNx    \undefined \def \showISBNx     #1{\unskip}     \fi
\ifx \showISBNxiii \undefined \def \showISBNxiii  #1{\unskip}     \fi
\ifx \showISSN     \undefined \def \showISSN      #1{\unskip}     \fi
\ifx \showLCCN     \undefined \def \showLCCN      #1{\unskip}     \fi
\ifx \shownote     \undefined \def \shownote      #1{#1}          \fi
\ifx \showarticletitle \undefined \def \showarticletitle #1{#1}   \fi
\ifx \showURL      \undefined \def \showURL       {\relax}        \fi
\providecommand\bibfield[2]{#2}
\providecommand\bibinfo[2]{#2}
\providecommand\natexlab[1]{#1}
\providecommand\showeprint[2][]{arXiv:#2}

\bibitem[Armanious et~al\mbox{.}(2019)]%
        {MedGAN_2019}
\bibfield{author}{\bibinfo{person}{Karim Armanious}, \bibinfo{person}{Chenming
  Jiang}, \bibinfo{person}{Sherif Abdulatif}, \bibinfo{person}{Thomas Kustner},
  \bibinfo{person}{Sergios Gatidis}, {and} \bibinfo{person}{Bin Yang}.}
  \bibinfo{year}{2019}\natexlab{}.
\newblock \showarticletitle{Unsupervised Medical Image Translation Using
  Cycle-{MedGAN}}. In \bibinfo{booktitle}{\emph{2019 27th European Signal
  Processing Conference ({EUSIPCO})}}. \bibinfo{publisher}{{IEEE}}.
\newblock
\urldef\tempurl%
\url{https://doi.org/10.23919/eusipco.2019.8902799}
\showDOI{\tempurl}


\bibitem[Ayd{\i}n and Karaarslan(2022)]%
        {aydin2022openai}
\bibfield{author}{\bibinfo{person}{{\"O}mer Ayd{\i}n} {and}
  \bibinfo{person}{Enis Karaarslan}.} \bibinfo{year}{2022}\natexlab{}.
\newblock \showarticletitle{OpenAI ChatGPT Generated Literature Review: Digital
  Twin in Healthcare}.
\newblock \bibinfo{journal}{\emph{Available at SSRN 4308687}}
  (\bibinfo{year}{2022}).
\newblock


\bibitem[Brunner et~al\mbox{.}(2018)]%
        {MusicGAN_2018}
\bibfield{author}{\bibinfo{person}{Gino Brunner}, \bibinfo{person}{Yuyi Wang},
  \bibinfo{person}{Roger Wattenhofer}, {and} \bibinfo{person}{Sumu Zhao}.}
  \bibinfo{year}{2018}\natexlab{}.
\newblock \bibinfo{title}{Symbolic Music Genre Transfer with CycleGAN}.
\newblock
\newblock
\urldef\tempurl%
\url{https://doi.org/10.48550/ARXIV.1809.07575}
\showDOI{\tempurl}


\bibitem[Chang et~al\mbox{.}(2018)]%
        {Character_2018}
\bibfield{author}{\bibinfo{person}{Bo Chang}, \bibinfo{person}{Qiong Zhang},
  \bibinfo{person}{Shenyi Pan}, {and} \bibinfo{person}{Lili Meng}.}
  \bibinfo{year}{2018}\natexlab{}.
\newblock \showarticletitle{Generating Handwritten Chinese Characters Using
  {CycleGAN}}. In \bibinfo{booktitle}{\emph{2018 {IEEE} Winter Conference on
  Applications of Computer Vision ({WACV})}}. \bibinfo{publisher}{{IEEE}}.
\newblock
\urldef\tempurl%
\url{https://doi.org/10.1109/wacv.2018.00028}
\showDOI{\tempurl}


\bibitem[Deng et~al\mbox{.}(2009)]%
        {imagenet_2009}
\bibfield{author}{\bibinfo{person}{Jia Deng}, \bibinfo{person}{Wei Dong},
  \bibinfo{person}{Richard Socher}, \bibinfo{person}{Li-Jia Li},
  \bibinfo{person}{Kai Li}, {and} \bibinfo{person}{Li Fei-Fei}.}
  \bibinfo{year}{2009}\natexlab{}.
\newblock \showarticletitle{ImageNet: A large-scale hierarchical image
  database}. In \bibinfo{booktitle}{\emph{2009 IEEE Conference on Computer
  Vision and Pattern Recognition}}. \bibinfo{pages}{248--255}.
\newblock
\urldef\tempurl%
\url{https://doi.org/10.1109/CVPR.2009.5206848}
\showDOI{\tempurl}


\bibitem[Ding et~al\mbox{.}(2021)]%
        {FPD_2021}
\bibfield{author}{\bibinfo{person}{Yifan Ding}, \bibinfo{person}{Liqiang Wang},
  {and} \bibinfo{person}{Boqing Gong}.} \bibinfo{year}{2021}\natexlab{}.
\newblock \showarticletitle{Analyzing Deep Neural Network’s Transferability
  via Fréchet Distance}. In \bibinfo{booktitle}{\emph{2021 IEEE Winter
  Conference on Applications of Computer Vision (WACV)}}.
  \bibinfo{pages}{3931--3940}.
\newblock
\urldef\tempurl%
\url{https://doi.org/10.1109/WACV48630.2021.00398}
\showDOI{\tempurl}


\bibitem[Fei-Fei et~al\mbox{.}(2006)]%
        {fei2006one}
\bibfield{author}{\bibinfo{person}{Li Fei-Fei}, \bibinfo{person}{Robert
  Fergus}, {and} \bibinfo{person}{Pietro Perona}.}
  \bibinfo{year}{2006}\natexlab{}.
\newblock \showarticletitle{One-shot learning of object categories}.
\newblock \bibinfo{journal}{\emph{IEEE transactions on pattern analysis and
  machine intelligence}} \bibinfo{volume}{28}, \bibinfo{number}{4}
  (\bibinfo{year}{2006}), \bibinfo{pages}{594--611}.
\newblock


\bibitem[Glorot and Bengio(2010)]%
        {normal_init_2010}
\bibfield{author}{\bibinfo{person}{Xavier Glorot} {and} \bibinfo{person}{Yoshua
  Bengio}.} \bibinfo{year}{2010}\natexlab{}.
\newblock \showarticletitle{Understanding the difficulty of training deep
  feedforward neural networks}. In \bibinfo{booktitle}{\emph{Proceedings of the
  Thirteenth International Conference on Artificial Intelligence and
  Statistics}} \emph{(\bibinfo{series}{Proceedings of Machine Learning
  Research}, Vol.~\bibinfo{volume}{9})},
  \bibfield{editor}{\bibinfo{person}{Yee~Whye Teh} {and} \bibinfo{person}{Mike
  Titterington}} (Eds.). \bibinfo{publisher}{PMLR}, \bibinfo{address}{Chia
  Laguna Resort, Sardinia, Italy}, \bibinfo{pages}{249--256}.
\newblock
\urldef\tempurl%
\url{https://proceedings.mlr.press/v9/glorot10a.html}
\showURL{%
\tempurl}


\bibitem[Goodfellow et~al\mbox{.}(2014)]%
        {GAN_2014}
\bibfield{author}{\bibinfo{person}{Ian~J. Goodfellow}, \bibinfo{person}{Jean
  Pouget-Abadie}, \bibinfo{person}{Mehdi Mirza}, \bibinfo{person}{Bing Xu},
  \bibinfo{person}{David Warde-Farley}, \bibinfo{person}{Sherjil Ozair},
  \bibinfo{person}{Aaron Courville}, {and} \bibinfo{person}{Yoshua Bengio}.}
  \bibinfo{year}{2014}\natexlab{}.
\newblock \bibinfo{title}{Generative Adversarial Networks}.
\newblock
\newblock
\urldef\tempurl%
\url{https://doi.org/10.48550/ARXIV.1406.2661}
\showDOI{\tempurl}


\bibitem[Guo et~al\mbox{.}(2019)]%
        {guo2019spottune}
\bibfield{author}{\bibinfo{person}{Yunhui Guo}, \bibinfo{person}{Honghui Shi},
  \bibinfo{person}{Abhishek Kumar}, \bibinfo{person}{Kristen Grauman},
  \bibinfo{person}{Tajana Rosing}, {and} \bibinfo{person}{Rogerio Feris}.}
  \bibinfo{year}{2019}\natexlab{}.
\newblock \showarticletitle{Spottune: transfer learning through adaptive
  fine-tuning}. In \bibinfo{booktitle}{\emph{Proceedings of the IEEE/CVF
  conference on computer vision and pattern recognition}}.
  \bibinfo{pages}{4805--4814}.
\newblock


\bibitem[He et~al\mbox{.}(2016)]%
        {resnet_2016}
\bibfield{author}{\bibinfo{person}{Kaiming He}, \bibinfo{person}{Xiangyu
  Zhang}, \bibinfo{person}{Shaoqing Ren}, {and} \bibinfo{person}{Jian Sun}.}
  \bibinfo{year}{2016}\natexlab{}.
\newblock \showarticletitle{Deep Residual Learning for Image Recognition}. In
  \bibinfo{booktitle}{\emph{2016 IEEE Conference on Computer Vision and Pattern
  Recognition (CVPR)}}. \bibinfo{pages}{770--778}.
\newblock
\urldef\tempurl%
\url{https://doi.org/10.1109/CVPR.2016.90}
\showDOI{\tempurl}


\bibitem[Heusel et~al\mbox{.}(2017)]%
        {FID_2017}
\bibfield{author}{\bibinfo{person}{Martin Heusel}, \bibinfo{person}{Hubert
  Ramsauer}, \bibinfo{person}{Thomas Unterthiner}, \bibinfo{person}{Bernhard
  Nessler}, {and} \bibinfo{person}{Sepp Hochreiter}.}
  \bibinfo{year}{2017}\natexlab{}.
\newblock \showarticletitle{GANs Trained by a Two Time-Scale Update Rule
  Converge to a Local Nash Equilibrium}.
\newblock  (\bibinfo{year}{2017}).
\newblock
\urldef\tempurl%
\url{https://doi.org/10.48550/ARXIV.1706.08500}
\showDOI{\tempurl}


\bibitem[Huang et~al\mbox{.}(2016)]%
        {huang2016deep}
\bibfield{author}{\bibinfo{person}{Gao Huang}, \bibinfo{person}{Yu Sun},
  \bibinfo{person}{Zhuang Liu}, \bibinfo{person}{Daniel Sedra}, {and}
  \bibinfo{person}{Kilian~Q Weinberger}.} \bibinfo{year}{2016}\natexlab{}.
\newblock \showarticletitle{Deep networks with stochastic depth}. In
  \bibinfo{booktitle}{\emph{Computer Vision--ECCV 2016: 14th European
  Conference, Amsterdam, The Netherlands, October 11--14, 2016, Proceedings,
  Part IV 14}}. Springer, \bibinfo{pages}{646--661}.
\newblock


\bibitem[Isola et~al\mbox{.}(2016)]%
        {I2I_2016}
\bibfield{author}{\bibinfo{person}{Phillip Isola}, \bibinfo{person}{Jun-Yan
  Zhu}, \bibinfo{person}{Tinghui Zhou}, {and} \bibinfo{person}{Alexei~A.
  Efros}.} \bibinfo{year}{2016}\natexlab{}.
\newblock \bibinfo{title}{Image-to-Image Translation with Conditional
  Adversarial Networks}.
\newblock
\newblock
\urldef\tempurl%
\url{https://doi.org/10.48550/ARXIV.1611.07004}
\showDOI{\tempurl}


\bibitem[Johnson et~al\mbox{.}(2016)]%
        {Johnson2016Perceptual}
\bibfield{author}{\bibinfo{person}{Justin Johnson}, \bibinfo{person}{Alexandre
  Alahi}, {and} \bibinfo{person}{Li Fei-Fei}.} \bibinfo{year}{2016}\natexlab{}.
\newblock \showarticletitle{Perceptual losses for real-time style transfer and
  super-resolution}. In \bibinfo{booktitle}{\emph{European Conference on
  Computer Vision}}.
\newblock


\bibitem[Kingma and Ba(2014)]%
        {adam_2014}
\bibfield{author}{\bibinfo{person}{Diederik~P. Kingma} {and}
  \bibinfo{person}{Jimmy Ba}.} \bibinfo{year}{2014}\natexlab{}.
\newblock \bibinfo{title}{Adam: A Method for Stochastic Optimization}.
\newblock
\newblock
\urldef\tempurl%
\url{https://doi.org/10.48550/ARXIV.1412.6980}
\showDOI{\tempurl}


\bibitem[Lake et~al\mbox{.}(2011)]%
        {lake2011one}
\bibfield{author}{\bibinfo{person}{Brenden Lake}, \bibinfo{person}{Ruslan
  Salakhutdinov}, \bibinfo{person}{Jason Gross}, {and} \bibinfo{person}{Joshua
  Tenenbaum}.} \bibinfo{year}{2011}\natexlab{}.
\newblock \showarticletitle{One shot learning of simple visual concepts}. In
  \bibinfo{booktitle}{\emph{Proceedings of the annual meeting of the cognitive
  science society}}, Vol.~\bibinfo{volume}{33}.
\newblock


\bibitem[Meng et~al\mbox{.}(2019)]%
        {meng2019glyce}
\bibfield{author}{\bibinfo{person}{Yuxian Meng}, \bibinfo{person}{Wei Wu},
  \bibinfo{person}{Fei Wang}, \bibinfo{person}{Xiaoya Li},
  \bibinfo{person}{Ping Nie}, \bibinfo{person}{Fan Yin}, \bibinfo{person}{Muyu
  Li}, \bibinfo{person}{Qinghong Han}, \bibinfo{person}{Xiaofei Sun}, {and}
  \bibinfo{person}{Jiwei Li}.} \bibinfo{year}{2019}\natexlab{}.
\newblock \showarticletitle{Glyce: Glyph-vectors for chinese character
  representations}.
\newblock \bibinfo{journal}{\emph{Advances in Neural Information Processing
  Systems}}  \bibinfo{volume}{32} (\bibinfo{year}{2019}).
\newblock


\bibitem[Pataranutaporn et~al\mbox{.}(2021)]%
        {pataranutaporn2021ai}
\bibfield{author}{\bibinfo{person}{Pat Pataranutaporn},
  \bibinfo{person}{Valdemar Danry}, \bibinfo{person}{Joanne Leong},
  \bibinfo{person}{Parinya Punpongsanon}, \bibinfo{person}{Dan Novy},
  \bibinfo{person}{Pattie Maes}, {and} \bibinfo{person}{Misha Sra}.}
  \bibinfo{year}{2021}\natexlab{}.
\newblock \showarticletitle{AI-generated characters for supporting personalized
  learning and well-being}.
\newblock \bibinfo{journal}{\emph{Nature Machine Intelligence}}
  \bibinfo{volume}{3}, \bibinfo{number}{12} (\bibinfo{year}{2021}),
  \bibinfo{pages}{1013--1022}.
\newblock


\bibitem[Pengcheng et~al\mbox{.}(2017)]%
        {pengcheng2017chinese}
\bibfield{author}{\bibinfo{person}{Gao Pengcheng}, \bibinfo{person}{Gu Gang},
  \bibinfo{person}{Wu Jiangqin}, {and} \bibinfo{person}{Wei Baogang}.}
  \bibinfo{year}{2017}\natexlab{}.
\newblock \showarticletitle{Chinese calligraphic style representation for
  recognition}.
\newblock \bibinfo{journal}{\emph{International Journal on Document Analysis
  and Recognition (IJDAR)}}  \bibinfo{volume}{20} (\bibinfo{year}{2017}),
  \bibinfo{pages}{59--68}.
\newblock


\bibitem[Pinder-Wilson et~al\mbox{.}(2023)]%
        {pinder-wilson_barbour_williams_2023}
\bibfield{author}{\bibinfo{person}{Ralph~H. Pinder-Wilson},
  \bibinfo{person}{Ruth Barbour}, {and} \bibinfo{person}{Robert Williams}.}
  \bibinfo{year}{2023}\natexlab{}.
\newblock \bibinfo{title}{Calligraphy}.
\newblock
\newblock
\urldef\tempurl%
\url{https://www.britannica.com/art/calligraphy}
\showURL{%
\tempurl}


\bibitem[Radford et~al\mbox{.}(2019)]%
        {radford2019language}
\bibfield{author}{\bibinfo{person}{Alec Radford}, \bibinfo{person}{Jeff Wu},
  \bibinfo{person}{Rewon Child}, \bibinfo{person}{David Luan},
  \bibinfo{person}{Dario Amodei}, {and} \bibinfo{person}{Ilya Sutskever}.}
  \bibinfo{year}{2019}\natexlab{}.
\newblock \showarticletitle{Language Models are Unsupervised Multitask
  Learners}.
\newblock \bibinfo{journal}{\emph{OpenAI}} (\bibinfo{year}{2019}).
\newblock
\urldef\tempurl%
\url{https://cdn.openai.com/better-language-models/language_models_are_unsupervised_multitask_learners.pdf}
\showURL{%
\tempurl}


\bibitem[Rezende et~al\mbox{.}(2017)]%
        {rezende2017malicious}
\bibfield{author}{\bibinfo{person}{Edmar Rezende}, \bibinfo{person}{Guilherme
  Ruppert}, \bibinfo{person}{Tiago Carvalho}, \bibinfo{person}{Fabio Ramos},
  {and} \bibinfo{person}{Paulo De~Geus}.} \bibinfo{year}{2017}\natexlab{}.
\newblock \showarticletitle{Malicious software classification using transfer
  learning of resnet-50 deep neural network}. In \bibinfo{booktitle}{\emph{2017
  16th IEEE International Conference on Machine Learning and Applications
  (ICMLA)}}. IEEE, \bibinfo{pages}{1011--1014}.
\newblock


\bibitem[Roemmele(2016)]%
        {roemmele2016writing}
\bibfield{author}{\bibinfo{person}{Melissa Roemmele}.}
  \bibinfo{year}{2016}\natexlab{}.
\newblock \showarticletitle{Writing stories with help from recurrent neural
  networks}. In \bibinfo{booktitle}{\emph{Proceedings of the AAAI Conference on
  Artificial Intelligence}}, Vol.~\bibinfo{volume}{30}.
\newblock


\bibitem[Rombach et~al\mbox{.}(2021)]%
        {SD_2021}
\bibfield{author}{\bibinfo{person}{Robin Rombach}, \bibinfo{person}{Andreas
  Blattmann}, \bibinfo{person}{Dominik Lorenz}, \bibinfo{person}{Patrick
  Esser}, {and} \bibinfo{person}{Björn Ommer}.}
  \bibinfo{year}{2021}\natexlab{}.
\newblock \bibinfo{title}{High-Resolution Image Synthesis with Latent Diffusion
  Models}.
\newblock
\newblock
\showeprint[arxiv]{2112.10752}~[cs.CV]


\bibitem[Shi et~al\mbox{.}(2015)]%
        {shi2015radical}
\bibfield{author}{\bibinfo{person}{Xinlei Shi}, \bibinfo{person}{Junjie Zhai},
  \bibinfo{person}{Xudong Yang}, \bibinfo{person}{Zehua Xie}, {and}
  \bibinfo{person}{Chao Liu}.} \bibinfo{year}{2015}\natexlab{}.
\newblock \showarticletitle{Radical embedding: Delving deeper to chinese
  radicals}. In \bibinfo{booktitle}{\emph{Proceedings of the 53rd Annual
  Meeting of the Association for Computational Linguistics and the 7th
  International Joint Conference on Natural Language Processing (Volume 2:
  Short Papers)}}. \bibinfo{pages}{594--598}.
\newblock


\bibitem[Terzio{\u{g}}lu et~al\mbox{.}(2022)]%
        {terziouglu2022ad}
\bibfield{author}{\bibinfo{person}{S{\"u}meyra Terzio{\u{g}}lu},
  \bibinfo{person}{Kevser~Nur {\c{C}}o{\u{g}}alm{\i}{\c{s}}}, {and}
  \bibinfo{person}{Ahmet Bulut}.} \bibinfo{year}{2022}\natexlab{}.
\newblock \showarticletitle{Ad creative generation using reinforced generative
  adversarial network}.
\newblock \bibinfo{journal}{\emph{Electronic Commerce Research}}
  (\bibinfo{year}{2022}), \bibinfo{pages}{1--17}.
\newblock


\bibitem[Tuomi(2023)]%
        {tuomi2023ai}
\bibfield{author}{\bibinfo{person}{Aarni Tuomi}.}
  \bibinfo{year}{2023}\natexlab{}.
\newblock \showarticletitle{AI-Generated Content, Creative Freelance Work and
  Hospitality and Tourism Marketing}. In \bibinfo{booktitle}{\emph{Information
  and Communication Technologies in Tourism 2023: Proceedings of the ENTER 2023
  eTourism Conference, January 18-20, 2023}}. Springer,
  \bibinfo{pages}{323--328}.
\newblock


\bibitem[Wang(2020)]%
        {kaggle_2020}
\bibfield{author}{\bibinfo{person}{Yuanhao Wang}.}
  \bibinfo{year}{2020}\natexlab{}.
\newblock \bibinfo{title}{Chinese calligraphy styles by calligraphers}.
\newblock
\newblock
\urldef\tempurl%
\url{https://www.kaggle.com/datasets/yuanhaowang486/chinese-calligraphy-styles-by-calligraphers}
\showURL{%
\tempurl}


\bibitem[Wang et~al\mbox{.}(2003)]%
        {msssim_2003}
\bibfield{author}{\bibinfo{person}{Z. Wang}, \bibinfo{person}{E.P. Simoncelli},
  {and} \bibinfo{person}{A.C. Bovik}.} \bibinfo{year}{2003}\natexlab{}.
\newblock \showarticletitle{Multiscale structural similarity for image quality
  assessment}. In \bibinfo{booktitle}{\emph{The Thrity-Seventh Asilomar
  Conference on Signals, Systems \& Computers, 2003}},
  Vol.~\bibinfo{volume}{2}. \bibinfo{pages}{1398--1402 Vol.2}.
\newblock
\urldef\tempurl%
\url{https://doi.org/10.1109/ACSSC.2003.1292216}
\showDOI{\tempurl}


\bibitem[Xing and Feng(2016)]%
        {xing2016romanization}
\bibfield{author}{\bibinfo{person}{Huang Xing} {and} \bibinfo{person}{Xu
  Feng}.} \bibinfo{year}{2016}\natexlab{}.
\newblock \showarticletitle{The romanization of Chinese language}.
\newblock  (\bibinfo{year}{2016}).
\newblock


\bibitem[Zhu et~al\mbox{.}(2017)]%
        {CycleGAN_2017}
\bibfield{author}{\bibinfo{person}{Jun-Yan Zhu}, \bibinfo{person}{Taesung
  Park}, \bibinfo{person}{Phillip Isola}, {and} \bibinfo{person}{Alexei~A
  Efros}.} \bibinfo{year}{2017}\natexlab{}.
\newblock \showarticletitle{Unpaired Image-to-Image Translation using
  Cycle-Consistent Adversarial Networks}. In \bibinfo{booktitle}{\emph{Computer
  Vision (ICCV), 2017 IEEE International Conference on}}.
\newblock


\end{thebibliography}


\end{document}